\documentclass{article}
\usepackage{natbib}
\usepackage[T1]{fontenc}
\usepackage[utf8]{inputenc}
\usepackage{lmodern}

\usepackage[english]{babel}
\usepackage{csquotes}
\usepackage{endnotes} 
\usepackage{multicol}
\usepackage{tikz}
\usepackage{float}
\usepackage{hyperref}
\usepackage{tipa}
\usepackage{example-sentences}
\usepackage{xcolor}
\usepackage{caption}

\begin{document}
\title{Gender Bias in Machine Translation and The Era of Large Language Models}
\author{Eva Vanmassenhove}
\maketitle

\abstract{This chapter examines the role of Machine Translation in perpetuating gender bias, highlighting the challenges posed by cross-linguistic settings and statistical dependencies. A comprehensive overview of relevant existing work related to gender bias in both conventional Neural Machine Translation approaches and Generative Pretrained Transformer models employed as Machine Translation systems is provided. Through an experiment using ChatGPT (based on GPT-3.5) in an English-Italian translation context, we further assess ChatGPT's current capacity to address gender bias. The findings emphasize the ongoing need for advancements in mitigating bias in Machine Translation systems and underscore the importance of fostering fairness and inclusivity in language technologies.\\

}

\section{Introduction}

This chapter investigates the role of Machine Translation (MT) in perpetuating and shaping societal biases, with a specific emphasis on gender bias. Within the realm of MT, biases, specifically gender biases, are particularly prominent due to contrastive linguistic settings that necessitate disambiguation and explicitness in their representation of gender. The (over)reliance of MT systems on statistical patterns, can lead to erroneous disambiguation and thus the generation of morphologically incorrect alternatives \citep{vanmassenhove-etal-2021-machine}. Moreover disambiguation based solely on statistical likelihood combined with the fact that current (traditional) MT systems exhibit limited capability in providing multiple potential translation candidates reinforces or even exacerbates (potentially harmful) stereotypes \citep{cho-etal-2019-measuring,rescigno2020case, prates2020assessing}. We advocate for more accountability in the design, development, and deployment of MT systems to address these biases and limitations.

To explore the potential for addressing gender bias in an English-Italian translation setting, we investigate the extent to which ChatGPT (based on GPT-3.5), a powerful Large Language Model (LLM), handles gender across these languages and whether it can contribute and help with bias mitigation. Specifically, we examine whether ChatGPT (GPT-3.5) is capable of generating translations that effectively address gender bias when explicitly prompted to do so. The translations generated by the model, despite specific prompts regarding gender, seem to indicate that there are limitations to the model's ability to provide a systematic solution for gender bias in translation. While the overall observations aligned with our expectations and the related work, it was surprising that explicitly prompting for gender alternatives often led to additional biases and inaccurate translations rather than more balanced, fair and/or less-biased responses. These findings underscore the need for further exploration and development in this area. In light of the limitations and potential biases of current MT systems, we advocate for continued research and improvements to promote fair and unbiased translations. Given the multidisciplinary nature of the fields, we advocate for more hybridization both in terms of the technology by, for instance, integrating linguistic information into neural models as in the field itself through more collaboration among (computational) linguists, computer scientists, sociolinguists, ethicists and others.

The remainder of this chapter is organized as follows: Section \ref{sec:genderbias} provides an overview of gender bias in MT and highlights the linguistic settings and statistical dependencies that contribute to its manifestation. Section \ref{sec:relwork} presents related work and discusses existing efforts to address gender bias in MT systems. Section \ref{sec:exp} outlines the experiment, including the specific prompts and the analysis. Finally, Section \ref{sec:conclusion} concludes the article by summarizing our findings and their implications, emphasizing the need for further research and advancements in addressing gender bias in MT systems, and advocating for increased accountability in the development and deployment of these technologies. By conducting a systematic analysis of gender bias in MT and evaluating the potential of LLMs in mitigating such bias, we hope to contribute to the ongoing efforts in creating fair, unbiased, and inclusive language technologies.

\section{Gender Bias in MT}\label{sec:genderbias}

The advent of Neural Machine Translation (NMT) played a pivotal role in breaking down language barriers and facilitating knowledge exchange among diverse (linguistic) communities. With their success and adoption, it also became increasingly important to investigate and address the inherent limitations and shortcomings these systems possess. First, current systems, have a tendency to not only perpetuate but also exacerbate biases and lack control in order to handle contrastive linguistic challenges, such as gender, in a systematic manner \citep{vanmassenhove-etal-2019-lost,vanmassenhove-etal-2021-machine}. Second, these systems are unable to adhere and abide by language-specific guidelines or policies related to gender-inclusive or gender-neutral language. As the influence of MT systems grows, it becomes increasingly important to scrutinize and analyse the biases embedded within MT systems and their potential impact on societal inequalities.

One prevalent form of bias, which warrants particular attention in translation contexts, is gender bias. Within the context of MT, we define gender bias as an (explicit or implicit) skew across or within language(s) towards gender(s), leading to systematic associations with gender(s) and thus creating representational (and in some cases allocational) harms that perpetuate unfair, inaccurate and potentially discriminatory stereotypes.

Gender biases in MT are often exacerbated by the need for explicit representation and disambiguation of gender in contrastive linguistic settings. For example, the English sentence \textit{`my husband is a \textbf{kindergarten teacher [neutral]}'} contains a profession noun \textit{`kindergarten teacher'} that is not explicitly marked for gender. However, within this sentences, it is clear that it refers to \textit{my husband}, a masculine nouns with which it semantically agrees. The sentence is translated by Google Translate\endnote{Accessed on June 12, 2023.} as \textit{`mijn man is \textbf{kleuterjuf [feminine]}'} in Dutch, where \textit{`juf'} is a gender-marked noun that can be translated into English as \textit{`miss'}. In many other cases, the MT systems do not produce translation errors but consistently opt for gender stereotypical translations such as the French \textit{‘tu es \textbf{belle [feminine]}'} as a translation for the English gender-neutral \textit{‘you are \textbf{beautiful [neutral]}'} (where the English sentence is besides gender, also ambiguous in terms of number and degree of politeness as opposed to French) and \textit{‘tu es \textbf{intelligent} [masculine]'} as a translation for the English sentence \textit{‘you are \textbf{smart} [neutral]’}. Another illustration, based on a tweet by Dora Vargha \citeyearpar{Vargha2021}\endnote{Re-translated on June 12 2023 by the author.}, is presented in Table \ref{tbl:illustration1} with Hungarian as the source language and English as the target. Hungarian has a gender-neutral pronoun \textit{\"{o}} which can be translated into English as either \textit{he}, \textit{she} or gender-neutral \textit{they}. These translations were generated by Google Translate.

\begin{table}[h!]
\centering
\captionsetup{width=280pt}
\setlength{\tabcolsep}{15pt}
\begin{tabular}{ll}
        \textbf{Hungarian} & \textbf{English} \\
        \textbf{\"{O}} szép. & \textbf{She} is beautiful. \\
        \textbf{\"{O}} okos. & \textbf{He} is clever. \\
        \textbf{\"{O}} olvas. &  \textbf{He} reads. \\
        \textbf{\"{O}} varr. & \textbf{She} sews. \\
        \textbf{\"{O}} tanít. & \textbf{He} teaches. \\
         \textbf{\"{O}} főz. &  \textbf{She} cooks. \\
         \textbf{\"{O}} kutat. & \textbf{He} is researching. \\
         \textbf{\"{O}} gyereket nevel. & \textbf{She} is raising a child. \\
         \textbf{\"{O}} sok pénzt keres. & \textbf{He} makes a lot of money. \\
         \textbf{\"{O}} süteményt süt. & \textbf{She} bakes cakes.
        
    \end{tabular}
    \caption{Illustration Gender Bias in Hungarian-English translations provided by Google Translate}
    \label{tbl:illustration1}
\end{table}

While none of the generated translations in Table \ref{tbl:illustration1} are wrong from a linguistic point of view, they do exhibit a fair amount of bias and stereotyping when it comes to opting for the feminine or masculine pronoun in the automatically generated English translations. In a systematic examination of this gender bias phenomenon, \cite{prates2020assessing} translated sentences containing professions from 12 different gender-neutral language (including Hungarian) into English and examined how often female, male or gender-neutral pronouns appeared in the translations. Their analysis not only revealed a significant imbalance in the distribution of pronouns but also highlighted that this imbalance exceeds the gender disparities observed in society itself for fields such as STEM (Science, Technology, Engineering and Mathematics). Their comparison with results from the U.S. Bureau of Labor Statistics revealed that Google Translate underestimated the expected frequency of female translations, demonstrating an even higher prevalence of male defaults compared to what job occupation data alone would suggest.

\subsection{Gender Bias Sources} 

Different languages handle and mark gender in different ways, resulting in various types of potential ambiguities that are hard to resolve for MT systems, making a one-solution-fits-all very unlikely. The requirement for explicitation and disambiguation in combination with the fact that languages differ in terms of the absence or presence of gender features, the number of gender classes and whether and where gender features are explicitly marked sets gender aside form other biases in contrastive linguistic settings. By this, it not implied that other biases do not occur, however, they do manifest themselves in different, more implicit ways.\endnote{Although one could argue, that this implicitness makes it all the more dangerous.}

When state-of-the-art NMT systems are faced with (gender) ambiguous words in cross-linguistic settings, they will generate the most statistically probable translation derived from the training data. This would imply that, one of the main sources of bias in the output are the datasets used as an input, referred to as dataset bias. We want to emphasize that reducing the problem solely to a matter of data bias is an oversimplification since such a reduction neglects the intricate web of factors underlying these biases, including historical legacies and societal injustices \citep{birhane2021algorithmic}. However, given that an in-depth discussion of historical and societal injustices is out of the scope of this chapter, we will further zoom in on data bias where we distinguish between coverage and social bias.\endnote{For a more comprehensive exploration of this topic, we refer to \cite{birhane2021algorithmic} which provides valuable insights into the ethical dimensions of algorithmic systems and their impact while advocating for a more critical analysis in the field of machine learning and data science.} Additionally, we will briefly discuss statistical and algorithmic bias.

Coverage bias in the context of gender and MT, refers to the underrepresentation and/or the misrepresentation of gender identities and gender-related terminology in the data used for training. For instance, an analysis of \textit{Europarl} \citep{Koehn2005}, one of the more popular training datasets for MT systems, revealed a 2:1 male-female speaker ratio. Such an imbalanced dataset in terms of male/female representation, most likely implies a higher frequency of male pronouns, male-endings (nouns, adjectives...), increasing the likelihood of generating "male" translations when facing ambiguity. This can be interpreted in the broad way, since there is also evidence that word embeddings are conforming to an overall androcentric worldview \citep{petreski2023word}.

Taking the example below, a few possible translations for the English word \textit{`teacher' [neutral]} are provided in French: \textit{`enseignant' [masculine singular]}, \textit{`enseignante' [feminine]}, \textit{`enseignant-e' [binary alternative]} or \textit{`le personnel enseignant' [gender-neutral alternative]}.\endnote{We acknowledge that the gender-neutral alternative \textit{`le personnel enseignant'} is not the most literal translation of \textit{`teacher'} and furthermore implies multiple people (plural). However, in settings where \textit{`teacher'} is used in a generic way, it can be considered a valid gender-neutral alternative.}

 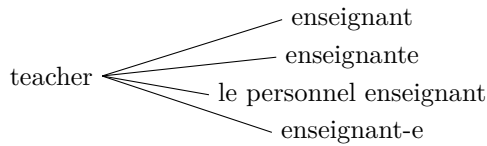
\begin{figure}[H]
\centering
\captionsetup{width=280pt}
\begin{tikzpicture}[node distance=2cm]
\node (teacher) [yshift=-0.5cm] {teacher};
\node (enseignant) [right of=teacher, xshift=2cm, yshift=0.75cm] {enseignant};
\node (enseignante) [right of=teacher, xshift=2cm, yshift=0.25cm] {enseignante};
\node (le personnel enseignant) [right of=teacher, xshift=2cm, yshift=-0.25cm] {le personnel enseignant};
\node (enseignant-e) [right of=teacher, xshift=2cm, yshift=-0.75cm] {enseignant-e};
\draw  (teacher.east) -- (enseignant.west);
\draw  (teacher.east) -- (enseignante.west);
\draw  (teacher.east) -- (enseignant-e.west);
\draw  (teacher.east) -- (le personnel enseignant.west);
\end{tikzpicture}
\caption{Some French translations of the English noun \textit{`teacher'}.}
\label{fig:teacher-translation}
\end{figure}

An imbalanced dataset could affect the probability of these translations being generated by an MT systems and affect the ability fo handle all possibilities. If the dataset contains a higher proportion of male speakers or masculine forms as opposed to feminine or gender-neutral ones, the MT system will become biased towards producing more masculine translations in case of ambiguity. It would thus consider \textit{`enseignant'} the more probable translation as opposed to the other valid options (given that there is no additional context to disambiguate). Additionally, a limited or no exposure to forms such as the binary gender inclusive \textit{`enseignant-e'} or the gender-neutral \textit{`le personnel enseignant'} in the training data can lead to erroneous or suboptimal translations that are less fluent, accurate and lack naturalness. While balancing training data in terms of male and female genders can be a somewhat viable option in many cases to alleviate these (binary) biases, more and more languages are accommodating for non-binary gender\endnote{We will use the term \textit{non-binary gender} to refer to all genders that fall outside the binary female/male distinction.} employing different strategies (for instance, gender-neutral pronouns like \textit{they} in English, gender-neutral endings like \textit{x} in Spanish...), a trend that will need to be reflected in MT technology as well. Currently this is not the case, among others, due to a lack of readily available gender-neutral/inclusive data, posing additional challenges for these data-hungry models. Rule-based or hybrid approaches that leverage linguistic knowledge in order to automatically generate gender-neutral alternatives in a post-processing way could be further explored to address the data scarcity issue, especially given the rather systematic nature of gender agreement rules \citep{vanmassenhove2021neutral}. 

Social bias in MT, refers to the bias that arises due to societal stereotypes and norms surrounding gender roles, which are embedded in and (un)consciously transmitted via the language in the training data. Such biases may influence the system to generate translations that align with traditional gender stereotypes or assumptions. Even more so because of the use of word embeddings which are used to capture semantic relationships and associations between words based on patterns present in the data. These embeddings can inadvertently reinforce existing biases as they learn from the prevailing language usage, including gendered language and stereotypes and \textit{embed} co-occurence patterns in the representations of the word themselves. A recent study by ~\cite{caliskan2022gender} investigating the extent of gender bias in static word embeddings revealed that the most common words and concepts are strongly associated with men more so than with women. Additionally, their findings highlighted how various gender biases are still prevalent across multiple dimensions as they observed differences in terms of the part-of-speech (e.g. women are more often associated with adjectives), clusters of concepts (e.g. women are more associated with sexual content and men with big tech), and basic dimensions of the meaning of words (e.g. women are more associated with positive valence and men with dominance).

While data bias is a well-acknowledged phenomenon, from a technical point of view, also statistical and algorithmic bias introduced by the translation models are important dimensions of bias in MT since they can introduce, perpetuate and even exacerbate biases observed in the training data. In the context of MT, \cite{vanmassenhove-etal-2019-lost} investigated the effect of statistical biases on generated translations. A statistically biased MT system may exhibit preferences for frequently occurring words or sub-words, potentially overlooking less common but equally valid synonyms or morphological variants. This bias leads to a loss of morphological variety and can impact the system's ability to generate diverse and grammatically correct translations. \cite{toral2019post} reached similar conclusions focusing on automatic post-editing, demonstrating its simpler and more normalized nature compared to human post-edited or translated texts. Both these observations can, in turn, be linked directly to gender bias in translations. Looking back at Figure \ref{fig:teacher-translation}, it is likely that some of the potential translations are way more common in the data than others. The model would strengthen the more frequently occurring connections and \textit{weaken} those that are more rarely observed in the data. In a follow-up paper, \cite{vanmassenhove-etal-2021-machine} conducted experiments where they train a model with which they then translate the original training data. They give an example where the most commonly occurring form of the French lemma \textit{`pr\'{e}sident'} in the training data, the masculine singular form, becomes even more common after training. The least common word form of that same lemma, the feminine plural \textit{`pr\'{e}sidentes'}, decreases in frequency and is for some models even completely lost in translation. In line with the abovementioned research, recent work by \cite{costa2022interpreting} shows how different architectures trained on the same data exhibit different levels of gender bias. When comparing language-specific encoder-decoders to shared encoder-decoder models, the former are less biased \citep{costa2022interpreting}. If machine translationese, i.e. the language produced by MT systems,  represent simplified, less lexically and morphologically diverse versions of the training data, this also raises more general questions related to the sociolinguistic implications and its potential long-term effects on language \citep{vanmassenhove-etal-2021-machine}.

We ought to note, however, that while the abovementioned biases might explain the biased outputs observed in automatically generated translations, simply `debiasing' or 'balancing' the datasets or the model will not enable systems to provide multiple correct alternatives (e.g. the previously provided translations from Hungarian into English) since this would require the generation of multiple alternative translations. More balanced data might however enable systems to more accurately translate sentences where the gender of a referent is present (see the previously given example: \textit{`My husband is a kindergarten teacher'}) and improve the overall quality of translations for all genders. 

\section{Related Work}\label{sec:relwork}
While it is out of the scope of this chapter to discuss the entire body of research on gender bias in Natural Language Technology and MT, we will discuss the most pioneering, impactful and/or most closely related work. We will cover some of the existing research on gender bias in both neural MT systems (Section \ref{subsec:NMT}) and more recent work focusing on GPT models used as MT systems (Section \ref{subsec:LLM}). NMT models are typically bidirectional sequence-to-sequence (\textit{seq2seq})\citep{Sutskever2014} architectures with encoder-decoder frameworks while GPT (Generative Pre-trained Transformer) models are primary a language models for language generation tasks. Both the more commonly employed NMT systems (the type of model used by Google Translate\endnote{\url{https://translate.google.com/}}, DeepL\endnote{\url{https://www.deepl.com}}, and others.)  and GPT models are based on Transformers \citep{Vaswani2017-Transformer}. However, GPT models are decoder-only models as opposed to the usual encoder-decoder architectures of the conventional MT systems. They are furthermore primarily trained on monolingual data (as opposed to carefully curated parallel corpora) and they require a significantly larger number of parameters \citep{hendy2023good}. When using a GPT model for a translation task, it involves relying on the GPT model's decoding capabilities in order to generate target language translations given a specific input in a unidirectional autogressive way, generating one token at a time, conditioned on the source. 

\subsection{"Conventional" NMT systems}\label{subsec:NMT}
The related research on bias in MT can be largely divided into approaches that aim to mitigate bias by addressing the
(i) pre-processing stage (training data and/or representations); (ii) processing stage (algorithm); or (iii) post-processing
stage (output). A large body of research on gender bias in language technology focused on debiasing models by changing the internal representations of separate words (often limited to animate nouns or adjectives) as a pre-processing step (a.o. \cite{bolukbasi2016man, zhao-etal-2018-gender}). Debiasing is then usually done by identifying and capturing a gender subspace using seed words (such as \textit{‘she’}, \textit{‘he’}, \textit{‘girl’}, \textit{‘boy’}...) and subtracting the obtained gender dimension from words such as \textit{`plumber’}, \textit{`receptionist’} or \textit{`doctor’} that are gender-neutral and should not carry gender information. The approach has been criticised in subsequent papers due to the fact that there is no clear correlation between removing bias from the internal representations, the so-called intrinsic bias, and the bias observed in downstream tasks or the extrinsic bias \citep{gonen2019lipstick,goldfarb2021intrinsic}. Aside from the limited impact of these debiasing techniques on the actual translations \citep{Font2019}, pre-processing techniques have other shortcomings, for instance, these methods (i) rely on lists of manually selected seed words that are ought to capture the `gender’ dimension; (ii) are usually applied to animate nouns (and sometimes adjectives) while other part-of-speechs (for instance, verbs, pronouns) might also require debiasing; (iii) offer no control over the eventual output as the MT system will still offer only one translation option; and (iv) do not allow to go beyond the binary gender distinction, i.e. more recent phenomena such as gender-neutral they cannot be handled in an appropriate manner. 

Another type of preprocessing techniques that have been explored is the integration of additional features that can help resolve ambiguities from the source into the target language \citep{vanmassenhove2018getting, Moryossef2019}. \cite{vanmassenhove2018getting} built the first speaker-informed NMT system where a tag indicating the gender of the speaker was fed to the NMT system along with the source input. Although the gender tags (that were limited to binary gender) improved the translation in terms of automatic evaluation scores, the manual evaluation revealed some (potentially undesirable) side-effects. While the gender tags were intended to improve the morphological gender agreement with the gender of the speaker, they often resulted in different lexical word choices. Similarly \cite{Moryossef2019} employed a parataxis construction such as She told him:’ to help the translation system disambiguate the gender of referents. Both approaches are limited in terms of the referents that can be controlled (first person singular as a subject \citep{vanmassenhove2018getting}, and first person singular as a subject as well as the third person singular as the (in)direct object \citep{Moryossef2019}. In \cite{stafanovivcs2020mitigating}, a method is presented whereby the gender of the words in the target are projected onto the source. The grammatical gender annotations from the target language in the source data can be leveraged by the MT system. The idea is somewhat similar, although more fine-grained, to the previous approaches presented given that it is assumed that the additional information provided will be a strong enough learning signal for the MT system to generate alternative translation.
The evaluation of their approach is limited to automatic evaluation metrics based on the WinoMT test suite (Stanovsky et al. 2019). WinoMT is a limited test suite given that it contains only sentences that are unambiguous in the source (i.e. the gender of the referents can be inferred from the source). 

To the best of our knowledge, only one work has attempted to control gender by opening the NMT black-box. \cite{Bau2019} developed an unsupervised method to identify and (de)activate the neurons responsible for specific linguistic phenomena. From their experiments, it resulted that gender is the most difficult phenomenon to capture and control with a 21\% success rate using the best setup. Gender is particularly hard to control because the ‘gender neurons’ are distributed along the network making it a difficult property to modify. 

We would like to note that most of the related work focuses on gender from a binary point of view (male/female) with exception to the work by \cite{saunders-etal-2020-neural} who added gender-neutral tags to an existing dataset and replaced gendered endings by placeholders, and the work by \cite{vanmassenhove2021neutral} and \cite{sun2021they} which focuses on developing a gender-neutral rewriter for English.

For a more in-depth overview of concepts related to bias, a summary of work on gender bias and a discussion of mitigation strategies proposed (up until 2021) for MT, we refer the reader to the work by \cite{savoldi-etal-2021-gender}.

\subsection{GPT for MT}\label{subsec:LLM}

GPT models are a type of LLMs that started gaining mainstream popularity near the end of 2022, when ChatGPT's user-friendly interface\endnote{\url{https://chat.openai.com/}} provided easy access to some of the capabilities of GPT models, impressing the public by its human-like responses in a conversational setup \citep{Toews_2023}. GPT models are developed by OpenAI's\endnote{\url{www.openai.com}} research lab. The first one, GPT, was developed back in 2018. Since then, various newer versions saw the light of day (GPT-2 (2019), GPT-3 (2020) and GPT-4 (2023)..), which are all iterations of the same GPT language model. The main difference between the different versions being the (exponential) growth in parameters: from 117 million parameters (GPT) to 175 billion for GPT-3, making it at the time the largest LLM. Reuters\endnote{\url{https://www.reuters.com} article accessible via \url{https://shorturl.at/dewzZ}} reported that according to an UBS study, ChatGPT became the fastest growing consumer application ever recorded, reaching 100 million monthly active users in January 2023, only two months after its launch on November 30, 2022. Given its popularity and wide-range usage (recreational and professional), ChatGPT is considered the first notable threat to Google's monopoly\endnote{\url{https://www.nytimes.com} article accessible via {https://shorturl.at/ruRS3}}\citep{ghosh2023chatgpt}. 

When talking about LLMs, since 2023, it is common to distinguish between bi-directional Transformer architecture (for instance, Bidirectional Encoder Representations from Transformers (BERT) \citep{Devlin2019BERTPO}) and uni-directional ones such as the GPT models. While unidirectional language models process and generate strictly in one direction (they process input from left-to-right and generate output left-to-right, token per token), bidirectional ones can attend to both the left and right context when processing input and output sequences, conditioning on preceding and succeeding tokens to generate output. The choice between unidirectional and bidirectional language models depends on the specific requirements of the task at hand. Unidirectional models are well-suited for sequential text generation and are computationally efficient, while bidirectional models tend to excel in tasks that demand a deeper understanding of context and long-range dependencies within the text. Both type of models (e.g. BERT vs GPT-3) have been used for the task of MT, each with their own strengths and weaknesses. BERT itself is not typically used as a direct MT system, it has been incorporated into MT pipelines in various ways to improve translation performance and address certain challenges (see previous section). As for GPT models, the interactive nature of ChatGPT (based on GPT models) allows it to engage in back-and-forth conversations with users which is an interesting ability to experiment with in the context of bias since it implies we can explicitly prompt it to be mindful of biases or to generate multiple alternative translations and thus potentially mitigate biased output. This interactivity opens avenues for experimentation in addressing bias in MT and tailoring translations to meet user preferences and specific contexts. Hereafter, we will therefore focus on related work that specifically explores the use of GPT models for MT. We ought to note that many of these works present very recent and somtimes preliminary findings published on arxiv\endnote{\url{https://arxiv.org/}} which have been approved for posting after moderation but have not gone through a peer review process yet.

Various studies have already investigated the more general translation abilities of GPT models. Among those are the work by \cite{hendy2023good,jiao2023chatgpt} and \cite{wang2023document} which evaluates GPT models for MT on a sentence-level \citep{hendy2023good,jiao2023chatgpt,wang2023document} and on a document-level \cite{hendy2023good,wang2023document}. \cite{hendy2023good} conducted an extensive automatic and human evaluation. They assessed the performance of three GPT models (ChatGPT,
GPT3.5 (text-davinci-003), and text-davinci002) on 18 language pairs across 4 domains and compared it to state-of-the-art (research and commercial) translation systems. They used the top-ranked systems from the WMT evaluation campaign to identify the state-of-the-art and the respective baselines for comparison. \cite{jiao2023chatgpt} present a more preliminary study using ChatGPT (GPT-3) for MT. Their analysis is deemed preliminary since it contains only 50 sentences for every set/condition and is additionally limited by the fact that they only employ automatic evaluation metrics. Both studies, however, find that while GPT models can perform competitively with existing MT systems in high-resource\endnote{We ought to note that this is an ill-defined term.} settings/languages, they significantly underperform when used in low-resource setting or for distant language paris \citep{hendy2023good,jiao2023chatgpt}. \cite{jiao2023chatgpt} re-evaluated the performance with launch of GPT-4 (March 15, 2023) observing that the performance of ChatGPT based on GPT-4 became more comparable to commercial translation systems for more distant languages.\endnote{Similar to 'low-resource languages' the terms 'distant languages' is not very well-defined. In the paper itself, they refer to `European' vs `Asian' languages.} \cite{wang2023document} compare commercial MT systems (Google Translate\endnote{\url{https://translate.google.com/}}, DeepL\endnote{\url{https://www.deepl.com/en/translator}}, and Tencent TranSmart\endnote{\url{https://transmart.qq.com/}}) and document-level NMT models with GPT models and focus particularly on evaluating their discourse awareness through automatic and human evaluations. They conclude that ChatGPT outperforms the commercial MT systems according to human evaluation and is thus a potentially promising avenue to explore for document-level translation.

Rather than concentrating on a direct comparison between existing MT systems and GPT models, other research \citep{peng2023towards} has explored how to effectively harness the capabilities and potential of GPT models in the context of translation. These investigations seek to optimize and maximize the benefits that GPT models can offer for improving translation quality and addressing specific translation challenges, for instance by exploring different prompting strategies or the possibility for personalized MT.

In \cite{peng2023towards}, ChatGPT is evaluated and compared to Google Translate with an emphasis on how to maximize the potential of ChatGPT. Their findings highlight the importance of clarifying the task information. They find that ChatGPT performs better when you make the task at hand very clear by stating that you are expecting it to serve as an MT system in a particular domain. The temperature, a (hyper)parameter that can be set to control the level of randomness/unpredictibility in the texts it generates, has a relatively small impact when dealing with high-resource languages (i.e. languages for which we generally have a large amount of linguistic resources, e.g. German or French). Additionally, they conduct a 'chain of thought' experiment where ChatGPT is prompted to not only translate but also explain its `thoughts' while doing so. The chain-of-thought experiments, however, led to word-by-word translation, thus degradeding the overall translation quality. It is important to highlight here that their evaluation consisted only of automatic evaluation metrics (Comet \citep{rei2020comet} and SacreBLEU \citep{post-2018-call}. In contrast with the findings presented in \cite{peng2023towards},  chain-of-thought experiments \citep{wei2022chain} combined with error analysis \citep{lu2022toward}, an approach coined \textit{error analysis prompting}, can lead to human-like MT evaluation at both the system and segment level \citep{lu2023error}. When prompted to handle multiple translation in one query, they observed unstable scorings and biases. Aside from automatic evaluation metrics (COMET, BLEU \citep{Papineni2002}, BERTscore \citep{zhang2019bertscore} and BLEURT\citep{sellam2020bleurt}), they refer to a `human evaluation' which, in reality, consists of an evaluation based on a high-quality dataset annotated by humans (and thus not an actual evaluation conducted by humans). Their findings should be considered preliminary given that they sample only 10 segments from each group they investigate (groups are based on the number of tokens) for Chinese--English, English--German and their reliance on automatic evaluation metrics. 

Two other papers investigate the usefulness of additional linguistic information integrated in the prompts fed to ChatGPT. \cite{gu2023linguistically} investigate a two-step prompt strategy where they inject linguistic knowledge into the prompt designed to aid MT. In \cite{gao2023design}, aside from explicitly mentioning the translation task information (e.g. domain), POS-tags are injected. While these are potentially interesting avenues to explore from an academic point of view in order to explore GPT models' full potential as an MT system, we can question to what extent it is realistic to explore all possible prompts for various language pairs/settings and effectively expect MT users to employ those strategies.

While the aforementioned research focused on the quality and potential of GPT models as MT systems, they did not particularly look into its ability to handle specific cross-linguistic phenomenon or biases. In \cite{ghosh2023chatgpt}, the work most closely related to ours, they examine whether ChatGPT (based on GPT-3) perpetuates gender bias across Bengali and five other low-resource languages: Farsi, Malay, Tagalog, Thai and Turkish. All the investigated languages, aside from English, use gender-neutral pronouns. This implies that, when translating from one of those languages into English, the gender-neutral pronoun can be translated as either \textit{he} or \textit{she} (or gender-neutral \textit{they}). In this translation setup, it becomes apparent that ChatGPT perpetuates and amplifies biases related to gender roles and stereotypes, in terms of actions (e.g. \textit{`cooking'} is a female activity) and occupations (e.g. \textit{`doctor'} is a male occupation),  when translating gender-neutral pronouns into English. Aside from that, in the context of occupations, it confers more respect to men than to women.\endnote{Bengali has two pronouns, one of which is the more respectful form and they found it to be used mainly in sentences featuring a male pronoun.} Vice versa, when translating from English, the gender-neutral pronoun \textit{they} could easily be translated into the equivalent gender-neutral pronoun in the other languages, however, their experiments show that currently, ChatGPT is unable to do so. 

\cite{ghosh2023chatgpt} furthermore rightfully stress that, while other MT providers have been criticized for the biased outputs produced by their MT systems, OpenAI has claimed to have put extensive bias mitigation measures in place. Yet, the experiments seem to indicate that the same gender biases that have been observed and criticized in commercial systems (e.g. Google Translate or MS Translator\endnote{\url{https://www.microsoft.com/nl-nl/translator/}}) persist. Given the popularity of ChatGPT, this warrants further investigation and the authors vouch for a more human-centered approach. While our experiments are still preliminary, we aim to contribute to the investigation of gender biases in ChatGPT used for MT.

\section{Preliminary Experiment and Analysis}\label{sec:exp}
In this preliminary experiment, we explore how and to what extent ChatGPT can handle gender across the English-Italian Language pair. While English is a language with pronominal gender, where only the pronouns are gendered\endnote{A few exceptions are nouns such as \textit{'waitress'} vs \textit{'waiter'} or \textit{`actress'} vs \textit{`actor'}.}, Italian is a more highly inflected language where different categories (nouns, adjectives...) agree in gender with their referent. Additionally, as opposed to English, Italian is a language with grammatical gender meaning that also inanimate nouns are assigned to a gender category. The example below is one taken from the challenge set used for our experiment and it provides a translation for an English gender-neutral sentence into Italian, where the past participle \textit{`sentito/a/\textschwa [`to feel']} agrees with the gender of its subject \textit{you}.\endnote{We added a version that is not included in gENder-IT using the `\textschwa' which has been recommended by \cite{baiocco2023italian} as a non-gender-specific ending for Italian.}

\begin{multicols}{2}
\textbf{English}\\
\textit{`What did \textbf{you} expect it to feel like?}

\columnbreak
\textbf{Italian}\\
\textit{`Como pensavi che ti saresti \textbf{sentito}'} \\
\textit{`Como pensavi che ti saresti \textbf{sentita}'}
\textit{`Como pensavi che ti saresti \textbf{sentit\textschwa}'}
\end{multicols}

\subsection{Experimental Setup}
The following sections describe the experimental setup, datasets and promps along with the error analysis conducted. For the experiments, we accessed\endnote{We accessed ChatGPT to generated these translations between May 30, 2023 and June 5, 2023.} ChatGPT (based on GPT-3.5) through the API. Given that the related work on using GPT models for MT highlighted the importance of (explicit) prompting, we looked at two different prompting scenarios.

\subsubsection{Dataset}
We used a subset of an existing challenge dataset, gENder-IT \citep{vanmassenhove2021gender}, designed specifically for resolving natural gender phenomena in translation contexts. The challenge set is a word-level (human) annotated, adapted and cleaned version of a subset of the MuST-SHE corpus \citep{bentivogli2020}, but unlike MuST-SHE which is annotated on the paragraph level, the sentences have been split and annotated on the word-level with $<$F$>$ (female), $<$M$>$ (male) or $<$A$>$ (ambiguous) tags. It focuses on sentences containing textually ambiguous words (in terms of gender) in the English source and provides the possible translation alternatives (in terms of binary gender)\endnote{There is no crystallized approach to gender-neutral endings in Italian as to date.} on the Italian target side. Two example of tagged sentences from the gENder-IT corpus are provided below for illustration. The first one, Example \ref{ex:first}, only contains <F> tags since the gender of `ambiguous' words such as \textit{`I'} can be determined based on sentence context. Example \ref{ex:sec}, is one where within the sentence context, no assumption can or should be made with respect to the gender of the referent. To indicate that not all ambiguous words have the same referent, numbers are appended (e.g. $<$A1$>$, $<$A2$>$...). We used a subset of gENder-IT for the manual error analysis consisting for which we used the first 50 sentences of the gENderIT corpus.

\begin{examples}
\item\label{ex:first} \textit{`‘So she turned and she looked at her dad, and she said, “Dad, \textbf{I <F>} know how \textbf{you <M>} feel, but \textbf{I <F>} don’t believe in the death penalty.”’}
\item\label{ex:sec} \textit{`And it was there that another \textbf{nurse <A1>}, not the \textbf{nurse <A2>} \textbf{who <A2>} was looking after Mrs. \textbf{Drucker <F>} before, but another \textbf{nurse <A1>}, said three words to \textbf{me <A3>} that are the three words that most emergency \textbf{physicians <A4>} \textbf{I <A3>} know dread.'}
\end{examples}

\subsubsection{Prompting Scenarios}
GPT's autoregressive approach allows it to produce contextually relevant responses which allow for more engaging interactions or requests as opposed to the conventional MT systems. This combined with the fact that previous research highlighted the importance of explicit prompting, we conducted the experiments with two different prompting scenarios. Furthermore, given that the research by \cite{peng2023towards} showed an improvement in performance for MT when prepending information stating the task at hand, in both scenarios, we entered the following prompt before generating any translations: \textit{`You are a Machine translation system'}. We generated the translations one by one in both cases. In the first one, we simply prompted ChatGPT to translate sentences containing an ambiguous pronoun or noun into Italian. In the second one, we explicitly prompted it to provide all the possible alternatives in terms of gender.

\begin{examples}
\item \textbf{Prompt Scenario 1} \textit{`Can you translate the following sentence into Italian: + [insert English sentence]'}
\item \textbf{Prompt Scenario 2} \textit{`Can you translate the following sentence into Italian providing all the possible alternatives in terms of gender + [insert English sentence]'}
\end{examples}

Scenario 1 is probably closer to how an MT-user would prompt ChatGPT since they are not necessarily aware of the fact that the target language might differ from the source in terms of gender marking. It would also give us some insight into how gender is handled `by default'. Ideally, ChatGPT would provide alternative translation even when not explicitly prompted to do so. Scenario 2 explicitly asks for alternative translation in terms of gender. In this scenario we hope to explore what the potential of GPT models is with respect to handling gender phenomena across languages. We ought to note that the tags ($<$F$>$, $<$M$>$ and $<$A$>$) with which sentences are annotated in the gENder-IT challenge set were removed from the English sentences for the prompt. For convenience, we added the tags to the examples provided in the subsequent section.

\subsection{Manual error analysis}
We conducted a manual analysis/evaluation of the translations generated by ChatGPT for both scenarios.

\subsubsection{Prompt Scenario 1}
For Scenario 1, the one where the prompt did not explicitly state that alternatives are to be generated, only 5 out of the 50 sentences did not miss valid alternatives in terms of gender. Out of those 5, three times ChatGPT had provided both alternatives (see Example \ref{ex:1}, \textit{`voltato/a'}) in the translation and two times there were no missing alternatives since the translation differed from the original one in such a way that the translation did not contain any forms marking the gender (see Example  \ref{ex:1}).

\begin{examples}
\item\label{ex:1} \textbf{EN} \textit{``So I <A> turned around and I <A> looked at her, and I <A> said, "Are you <F> gonna wear this?"''}
\\ \textbf{IT} \textit{``Mi sono \textbf{voltato/a} e l'ho guardata dicendo: ""Lo indosserai?"''}

\item\label{ex:2} \textbf{EN} \textit{``What did you <A> expect it to feel like?''}
\\ \textbf{IT} \textit{``Cosa ti aspettavi che si sentisse?''}
\end{examples}

For all other 45 sentences, there were either errors or valid translation alternatives in terms of gender missing. Two sentences contained errors with respect to gender in the generated translation. In Example \ref{ex:3}, it is clear from the sentential context that \textit{`classmate'} refers to someone with a female gender. Nevertheless, it is translated as \textit{`compagno [masculine]'} in the Italian translation generated by ChatGPT. Similarly, Example \ref{ex:4} shows how \textit{child} which should be translated into the feminine form according to the given context, is translated into Italian as \textit{`figlio'} which means \textit{`son'}. Note that in both cases, the error occurred due to the fact that a masculine form was generated while the correct form was feminine.

\begin{examples}
\item\label{ex:3} \textbf{EN} \textit{``A college \textbf{classmate <F>} of mine wrote me <A> a couple weeks ago and said she thought I <A> was a little strident.''}
\\ \textbf{IT} \textit{``Un \textbf{compagno} di classe del college mi ha scritto un paio di settimane fa dicendo che pensava che fossi un po' stridente.''}

\item\label{ex:4} \textbf{EN} \textit{``I <A1> knew that if in one hour, \textbf{our child <F>} didn't come, I <A1> was going to leave her there and go back to work and make a presentation about AI to my \textbf{ boss <A2>}, Apple's CEO <A2>.''}
\\ \textbf{IT} \textit{``Sapevo che se \textbf{nostro figlio} non fosse arrivato entro un'ora, l'avrei lasciata lì e sarei tornato al lavoro per fare una presentazione sull'IA \textbf{al mio capo}, il CEO di Apple.''}
\end{examples}

In the remaining 43 sentences there was at least one alternative translation missing. To illustrate that, we can once more refer back to Example \ref{ex:4} where \textit{`boss'} in English could be translated into Italian as \textit{capa [feminine]} or \textit{capo [masculine]}, yet only the masculine alternative was provided. When only considering the sentences with two alternative translations, in the vast majority of the cases, the feminine alternatives were lacking with the exception to a few sentences containing the noun `nurse', `secretary' and names stereotypically associates with the female gender (e.g. `Vera'). In the gENder-IT challenge set, however, they opted to not assign genders based solely on the names because of potential cultural biases - which all names are considered `ambiguous' unless a full name is provided and the gender of the person could be retrieved through a Wikipedia page (i.e. public/historical figures).

\subsubsection{Prompt Scenario 2}

For Scenario 2, ChatGPT was prompted to translate the same 50 sentences, the only difference being that this time it was explicitly stated that it should generate all possible alternatives in terms of gender for the translation. Out of the 50 sentences, 16 of the translations were entirely correct with no missing alternatives. For half of the sentences, valid alternatives were missing and in 17 sentences there were erroneous translations (in terms of gender). In this scenario, all the alternatives missing were alternatives containing feminine forms. For instance, in \ref{ex:5} only one translation is generated while \textit{`pediatrician'} is ambiguous in the source context and could be translated into the feminine or masculine form in Italian (\textit{dal pediatra [masculine]} or \textit{dalla pediatra [feminine]}) yet only the masculine alternative is provided. Note as well how the proper noun `Sarah' is translated as `Laura'.

\begin{examples}
\item\label{ex:5} \textbf{EN} \textit{``And when Elliot was first born, we <A1> took her to the \textbf{pediatrician <A2>} and we met our first nurse <F> — we'll call her Sarah.''}
\\ \textbf{IT} \textit{``E quando Elliot è nata per la prima volta, l'abbiamo portata \textbf{dal pediatra} e abbiamo incontrato la nostra prima infermiera — la chiameremo Laura.''}
\end{examples}

Arguably more problematic than the missing alternatives, were the alternatives that were erroneous since they did not stay true to the source. In Example \ref{ex:7}, four alternative translations were generated given the English source sentence. While the first alternative stays true to the source by translating \textit{`mother'} into \textit{`madre'}, in the subsequent alternatives it is erroneously translated into the Italian as \textit{`padre'}, \textit{`fratello'} and \textit{`sorella'} which mean \textit{`father'}, \textit{`brother'} and \textit{`sister'} in English.

\begin{examples}
\item\label{ex:7} \textbf{EN} \textit{``My \textbf{mother} had been left behind in the camp, and I <A1> had not seen her since I <A1> left it <A2>.''}
\\ \textbf{IT}: Alternative 1 \textit{``Mia \textbf{madre} era stata lasciata indietro nel campo e non l'avevo più vista da quando l'avevo lasciato.''}
\\ \textbf{IT}: Alternative 2  \textit{``Mio \textbf{padre} era stato lasciato indietro nel campo e non l'avevo più visto da quando l'avevo lasciato.''}
\\ \textbf{IT}: Alternative 3  \textit{``Mio \textbf{fratello} era stato lasciato indietro nel campo e non l'avevo più visto da quando l'avevo lasciato.''}
\\ \textbf{IT}: Alternative 4  \textit{``Mia \textbf{sorella} era stata lasciata indietro nel campo e non l'avevo più vista da quando l'avevo lasciata.''}

\end{examples}

Similarly, in Example \ref{ex:8}, ChatGPT manages to correctly provide two alternatives for the gender-neutral English \textit{psychiatrist}, \textit{un/una psichiatra}. However, it then proceeds by providing two additional inaccurate alternatives where \textit{psychiatrist} is translated as \textit{ginecologo} (\textit{gynaecologist}).

\begin{examples}
\item\label{ex:8} \textbf{EN} \textit{``I'm a psychiatrist <A> who <A> works with pregnant and postpartum women...''}
\\ \textbf{IT}: Alternative 1 \textit{``Sono \textbf{uno psichiatra} che lavora con donne in gravidanza e nel periodo postpartum...''}
\\ \textbf{IT}: Alternative 2  \textit{``Sono \textbf{una psichiatra} che lavora con donne in gravidanza e nel periodo postpartum...''}
\\ \textbf{IT}: Alternative 3  \textit{``Sono \textbf{un ginecologo} che lavora con donne in gravidanza e nel periodo postpartum...''}
\\ \textbf{IT}: Alternative 4  \textit{``Sono \textbf{un ginecologo} che lavora con donne in gravidanza e nel periodo postpartum...''}
\end{examples}

\subsection{Findings \& Implications}

Our preliminary findings on the small challenge set are in line with the findings of \cite{ghosh2023chatgpt}. Our analysis, of both prompting scenarios, indicates a (strong) male bias which seems to become even more prevalent when asked explicitly to generate alternatives. While ChatGPT has some potential when it comes to generating alternatives when prompted explicitly, it currently cannot handle gender in a systematic manner. Given the current hype surrounding LLMs and their potential, we deem it important to raise awareness among researchers and potential users of these technologies. We will summarize the main findings regarding the model's handling of gender and its potential implications. 

First of all, in both the Scenario 1 and 2, the analysis revealed serious shortcomings in ChatGPT's ability to generate multiple possible translations in terms of gender. In Scenario 1, only a very small proportion of the translated sentences contained all binary gender translation alternatives. In Scenario 2, there were numerous instances where feminine gender alternatives were entirely missing. Furthermore, while in Italian, there is no crystallized approach yet to gender-neutral language, gender-neutral markers such as the asterisk or the schwa are used by the transgender Italian community. Such endings or markers were never provided or suggested by the model. Its inability to account for gender alternatives when not explicitly prompted to do so is already problematic as it demonstrates a lack of sensitivity to gender-inclusive language. However, the absence of gender-neutral and feminine alternatives, especially in Scenario 2, where explicit instructions were provided, reveals an actual inability of the model to handle gender systematically across languages. Both observations are leading to incomplete and biased translations and emphasize the need for ongoing improvements and vigilance. Furthermore, the absence of gender-neutral language markers in the model's output is a missed opportunity for fostering greater inclusivity. The failure to provide such alternatives reflects a lack of awareness of evolving language norms and the specific needs of different user groups.

Second, in Scenario 2, there were various erroneous translations among the alternatives provided. The model would provide alternatives that deviated from the source text and introduced misleading and incorrect information. These erroneous alternatives were sometimes not (directly) related to gender (Example \ref{ex:8}) but often the (feminine) terms were inaccurately replaced with masculine alternatives. These translations not only distort the intended message and thus the accuracy of the translation but also often perpetuate gender biases that are deeply ingrained in our society. Erroneous and biased translations that misrepresent the source content furthermore carry the risk of alienating or excluding individuals who do not conform to traditional gender norms.

Overall, these findings underscore the need for continued research and development to enhance our understanding of LLMs' handling of gender-related nuances in translation. Like more traditional MT models, ChatGPT struggles with gender (biases) across languages. The ability to explicitly prompt it to handle issues related to bias and its tendency to generate answers that at first sight seem to address the issue appropriately furthermore risks creating a feeling of complacency. It is however crucial to recognise that while LLMs, like GPT3.5 have shown remarkable capabilities in various NLP tasks, they are not immune to biases. While explicit prompting might be a useful technique to alleviate some biases, this does not guarantee consistent bias-free or fully accurate responses. When LLMs generate responses that seem to handle gender bias well on the surface, users might become complacent and assume that the issue has been adequately addressed while underlying biases persist. Vigilance is thus required to assess the generated content critically and carefully. Addressing these issues is not only a matter of linguistic accuracy but also a step towards promoting more inclusive and respectful communication in the digital age. The AI community at large, in collaboration with linguists, translators and a diverse set of users should collaborate to further evaluate these models to eventually minimize bias, and ensure that AI-generated content aligns with evolving societal standards of gender-inclusivity and neutrality.

\subsubsection{Limitations}
We would like to discuss some of the limitations of our work. First of all, there is a reproducibility problem that occurs when querying ChatGPT. The results of the same query/prompt may vary across multiple trials and the models underlying the app can be updated at any time. This implies that there is a certain randomness to the evaluation and analysis of the translations. Furthermore, currently, a novel LLM (GPT-4) has been developed and can be accessed through a \$20 monthly subscription to ChatGPT. According to \cite{jiao2023chatgpt}, GPT-4 outperforms GPT-3.5 as an MT system which warrants for a replication using GPT-4. Nevertheless, the free version of ChatGPT is still based on GPT-3.5, implying that a large amount of users will still be using the currently investigated model for their daily needs. 

Aside from the reproducibility limitations, the gENder-IT corpus does not contain any gender-neutral sentences (e.g. using the gender-neutral pronoun \textit{they} in the English source). We prompted ChatGPT to generate gender alternatives but did not ask specifically for gender-neutral alternatives. While \cite{ghosh2023chatgpt} finds that ChatGPT cannot handle the English gender-neutral pronoun, they also did not explicitly prompt it to translate in a gender-neutral way.

\section{Conclusion}\label{sec:conclusion}

In conclusion, this chapter has delved into the critical role of MT in perpetuating and shaping societal biases, particularly gender bias. Within the realm of MT, gender biases are especially pronounced due to the need for disambiguation and explicit representation of gender in specific contrastive linguistic settings. The reliance on biased data and statistical patterns in current MT systems can lead to erroneous disambiguation, resulting in the generation of morphologically incorrect alternatives, reinforcing potentially harmful stereotypes.

We provided an overview of existing work related to gender bias in both conventional neural machine translation (NMT) approaches and the use of GPT models as MT systems. We furthermore covered and explained some of the origins/sources of gender bias we observe in language technology. Given the rising popularity of tools like ChatGPT, it is crucial to raise awareness about these issues and take proactive steps to address them. Our experiments with ChatGPT, based on GPT-3.5, in an English-Italian translation setting showed how GPT models perpetuate biases even when explicitly prompted to provide alternative translation. Surprisingly, prompting for gender alternatives often even led to additional biases in the model's outputs, highlighting the need for further research and development in this area.

The implications of these findings underscore the importance of ongoing research and improvements to promote fair and unbiased translations. We advocate for more hybridization in both technology and collaboration among researchers from various disciplines, including computational linguists, computer scientists, sociolinguists, and ethicists. Such multidisciplinary efforts are essential to address the complex and multifaceted nature of gender bias in MT. We acknowledge that our experiments have been conducted with a relatively small dataset. To gain a deeper understanding of GPT's ability to deal with gender across languages and to expand the scope of our research, a more detailed analysis with a larger set of sentences across multiple language pairs should be conducted in future work to assess the model's effectiveness across different linguistic contexts.

Looking ahead, an important area of future work is to investigate how GPT models can be used to generate gender-inclusive datasets. This involves exploring ways to integrate linguistic knowledge into the model to guide the generation of gender alternatives, including those beyond the binary. One potential approach could be to provide tags, similar to those used in the gENder-IT challenge, to influence the model's output and encourage gender-inclusive translations. As the use of GPT models and other language generation technologies continues to grow, it becomes increasingly vital to ensure that these systems are equipped to handle gender and promote inclusivity. Our research aims to contribute to the ongoing efforts to make MT systems more equitable and sensitive to issues of gender representation.

In closing, this chapter has shed light on the challenges posed by gender bias in MT. It emphasizes the need for continued multidisciplinary research, advancements, and collaborative endeavors to pave the way towards more equitable and unbiased language technologies. As the field progresses, we hope to see the transformation of MT systems into powerful tools that promote diversity, inclusivity, and societal progress.

\section*{Acknowledgements} I would like to thank Esther Monz\'{o}-Nebot for setting up a collaborative initiative gathering expertise from linguistics, computer science and law in order to discuss some of the more pressing issues regarding translation technology. Additionally, I would like to thank Deborah Giustini, Heidi Salaets and Maurizio Veglio for sharing their expertise and broadening my knowledge across diverse fields.

\bibliography{main}
\theendnotes

\end{document}